\RequirePackage[OT1]{fontenc}
\documentclass[conference,letterpaper]{IEEEtran}
\IEEEoverridecommandlockouts
\usepackage{graphicx}
\usepackage{amsmath,amssymb}
\usepackage{booktabs}
\usepackage[numbers,sort&compress]{natbib}
\usepackage[dvipsnames]{xcolor}
\usepackage{newunicodechar}
\newunicodechar{’}{'}
\newunicodechar{•}{\ensuremath{\bullet}}
\usepackage{makecell}
\usepackage{multirow}
\usepackage{arydshln}
\usepackage{algorithm}
\usepackage{algorithmic}

\makeatletter
\let\originalIEEEcaption\@makecaption
\long\def\@makecaption#1#2{%
  \def\tabletype{table}%
  \ifx\@captype\tabletype
    \vskip\abovecaptionskip
    {\normalfont\small\raggedright Table~\thetable. #2\par}%
    \vskip 4pt
  \else
    \originalIEEEcaption{#1}{#2}%
  \fi}
\makeatother
\usepackage[hidelinks]{hyperref}
\title{SmoothShift: Disrupting Visual Adversarial Attacks on LVLMs at Test Time}
\author{\IEEEauthorblockN{Xiaoqing Chen$^{1,2,*}$, Jiayi Ouyang$^{1,*}$, Dongrui Wu\textsuperscript{1,2,\textdagger}}
\IEEEauthorblockA{$^{1}$Huazhong University of Science and Technology\\
$^{2}$Zhongguancun Academy}\thanks{\protect\parbox[t]{\dimexpr\columnwidth-1em\relax}{\raggedright
\textsuperscript{*}\ Equal contribution.\newline
\textsuperscript{\textdagger}\ Corresponding author.\newline
Emails: Xiaoqing Chen (xqchen914@hust.edu.cn), Dongrui Wu (drwu09@gmail.com).}}}

\begin{document}
\maketitle
\begin{abstract}
Visual adversarial perturbations can substantially impair large vision-language models (LVLMs) despite introducing only subtle input changes. To understand how these small changes cause substantial errors, we investigate their optimization and propagation through the model, revealing two key findings. First, stronger attacks rely on finer iterative optimization, while subtle input perturbations induce representation discrepancies that are amplified layer by layer through the vision encoder and persist in the language model. Second, disrupting the alignment between within-patch input patterns and the model's computation reduces clean--adversarial representation discrepancies and substantially weakens attacks. Together, these findings suggest that adversarial effectiveness depends on both fine-grained perturbation patterns and their coupling with the model's computation. Motivated by these findings, we propose SmoothShift, a lightweight, training-free defense that combines local smoothing and spatial transformations. Smoothing weakens fine-grained perturbation patterns, while spatial transformations disrupt their alignment with the encoder's patch grid. SmoothShift requires no model updates, auxiliary networks, or additional model inference. Experiments cover three LVLM families on four datasets and robustly fine-tuned Contrastive Language--Image Pre-training (CLIP) models on eight classification datasets. Evaluations across eight attack methods and multiple perturbation budgets demonstrate improved robustness while largely preserving clean performance. SmoothShift also provides robustness gains under defense-aware attacks using mixed-gradient BPDA with EOT.
\end{abstract}

\section{Introduction}

Large vision-language models (LVLMs) map an input image to a sequence of visual tokens. They process these tokens together with a textual query to generate open-ended responses~\cite{radford2021clip,li2023blip2,liu2023visual,awadalla2023openflamingo}. Despite their strong multimodal capabilities, LVLMs remain vulnerable to adversarial perturbations applied to the visual input. Previous studies have shown that small, visually imperceptible perturbations can substantially alter visual representations, impair cross-modal alignment, and induce incorrect or irrelevant responses~\cite{Schlarmann2023MMFM,cui2024robustness}.   More recent attacks improve effectiveness, transferability, and prompt generalization by targeting visual representations~\cite{Zhao2023AttackVLM,Wang2024VTAttack,Mei2026VEAttack}, optimizing across prompts or queries~\cite{Luo2024CroPA,Zhang2025QAVA}, or exploiting cross-modal semantics~\cite{Xie2025Chain}. These vulnerabilities undermine the trustworthiness of LVLMs, motivating research into defenses against visual adversarial attacks.

\begin{figure}
    \centering
    \includegraphics[width=\linewidth]{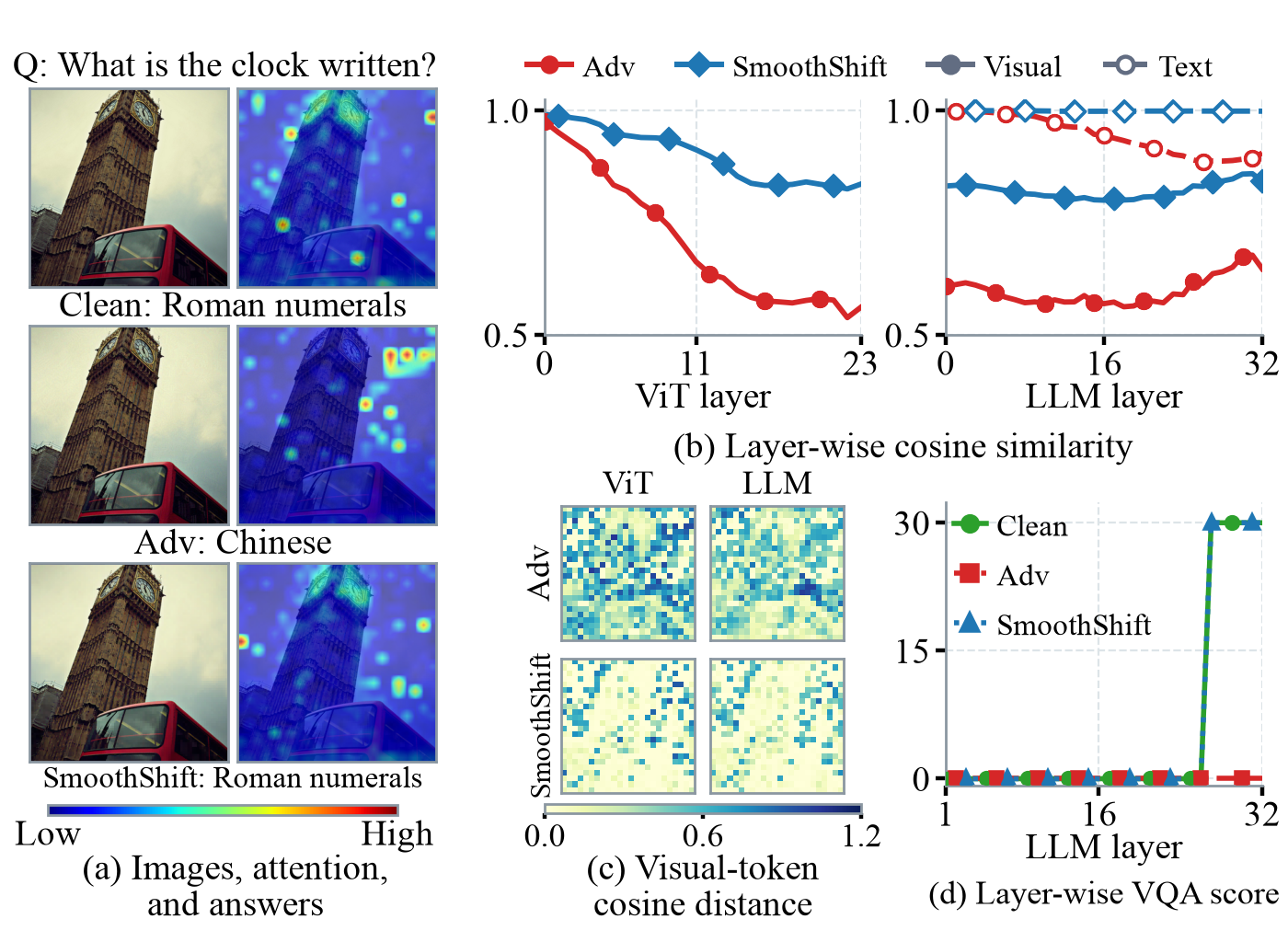}
\caption{Effects of SmoothShift on a VQA example using LLaVA-1.5-7B. (a) Images, attention maps, and generated answers. SmoothShift restores attention to relevant regions and recovers the correct answer. (b) Mean token-wise cosine similarity to clean-input representations across Vision Transformer (ViT) and large language model (LLM) layers. Solid and dashed curves denote visual and text tokens, respectively. (c) Visual-token cosine distances to clean-input representations at the ViT output and final LLM layer. (d) VQA scores for answers independently decoded from individual LLM layers.}
    \label{fig:method}
\end{figure}

Existing defenses against visual adversarial attacks on LVLMs broadly fall into training-based and test-time approaches. Training-based methods improve intrinsic robustness through adversarial adaptation of vision-language encoders~\cite{Mao2023,Schlarmann2024}, robust visual pre-training and multimodal tuning~\cite{WangZeyu2025DVD,malik2025robustllava}, or preference optimization~\cite{liu2026adpo}. Although effective, these methods incur considerable training costs and may also impair the models' generalization ability. Test-time methods mitigate adversarial effects through response correction~\cite{zhou2025dps}, image purification~\cite{zhang2025f3,Nie2022DiffPure}, prompt adaptation~\cite{WangXin2025TAPT}, or adversarial detection and adaptation~\cite{LiZhiwei2025TTP}. Together, these approaches provide effective ways to improve LVLM robustness, with test-time defenses offering deployment flexibility without retraining the underlying model. However, how optimized visual perturbations induce and sustain adversarial effects in LVLMs remains incompletely understood.

\begin{figure*}[t]
    \centering
    \includegraphics[width=\textwidth]{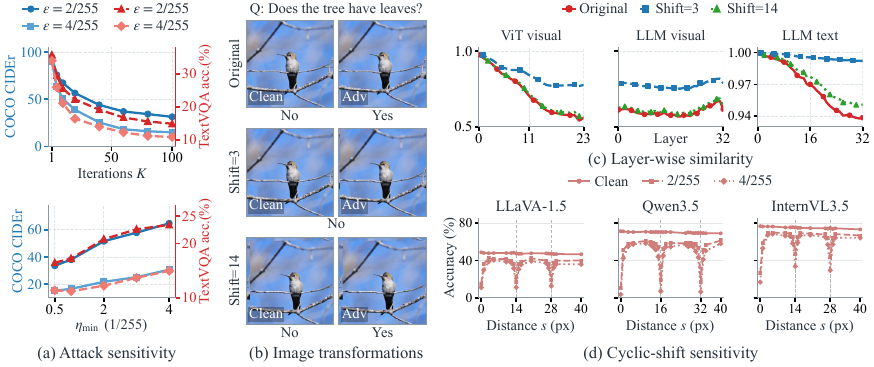}
    \caption{Optimization and spatial-alignment sensitivity of visual adversarial attacks. (a) More APGD iterations and smaller minimum step sizes strengthen attacks on COCO and TextVQA. (b) A three-pixel cyclic shift corrects the adversarial answer, whereas a fourteen-pixel shift does not. (c) Layer-wise cosine similarity between clean and adversarial representations on TextVQA. The three-pixel shift better preserves representation similarity. (d) TextVQA accuracy under cyclic shifts for LLaVA-1.5-7B, Qwen3.5-4B, and InternVL3.5-4B. Dashed vertical lines mark one and two patch widths, where mitigation weakens.}
    \label{fig:motivation}
\end{figure*}

To understand how subtle visual perturbations induce prediction errors, we begin by examining their optimization and propagation through the model. We observe that attacks become more effective with more iterations and finer step-size refinement~\cite{CroceHein2020AutoAttack}, as shown in Figure~\ref{fig:motivation}(a). Figure~\ref{fig:method}(a) shows that clean and adversarial images are visually similar. However, their representation discrepancies grow through successive layers of the vision encoder and persist in the language model, as shown in Figure~\ref{fig:method}(b). Figure~\ref{fig:method}(c) further illustrates these differences at the vision encoder's output and the final LLM layer. These deviations are accompanied by the incorrect answer in Figure~\ref{fig:method}(a) and the lower VQA score in Figure~\ref{fig:method}(d). These observations lead us to hypothesize that finely optimized perturbations exploit a precise coupling between the perturbed input and the model's computation. Rearranging input pixels offers a way to probe this coupling. Figure~\ref{fig:motivation}(d) shows that small cyclic shifts substantially weaken attacks, whereas shifts by one or two patch widths preserve within-patch alignment and provide much less mitigation. This contrast suggests that the coupling exploited by adversarial perturbations is sensitive to pixel arrangement within visual patches.

Building on these observations, we propose \textbf{SmoothShift}, a lightweight, training-free test-time defense that combines local smoothing with spatial transformations. Applied directly to the input image, SmoothShift requires no model updates, auxiliary networks, or additional LVLM forward passes. Smoothing disrupts finely optimized perturbation patterns, while spatial transformations alter the input's alignment with the vision encoder's spatial partition. Together, these operations aim to disrupt both the fine-grained structure of optimized perturbations and their coupling with the model's computation, while preserving task-relevant visual information. Figure~\ref{fig:method}(a) shows that SmoothShift restores attention to task-relevant image regions and recovers the correct answer. These improvements are accompanied by representations closer to their clean counterparts across the vision encoder and language model in Figure~\ref{fig:method}(b) and (c), and a restored VQA score in Figure~\ref{fig:method}(d).

Extensive experiments across four datasets, three LVLM families, eight attack methods, and four perturbation budgets demonstrate that SmoothShift improves adversarial performance on image captioning and visual question answering while largely preserving clean performance. Evaluations on robustly fine-tuned Contrastive Language--Image Pre-training (CLIP) models across eight classification datasets extend these benefits to image classification. We further examine defense-aware attacks and cross-configuration transfer, showing that SmoothShift continues to improve robustness over the undefended model even under stronger adaptive optimization.

Our contributions are summarized below.
\begin{itemize}
    \item We find that visual adversarial effectiveness depends on fine-grained optimization and within-patch pixel alignment. These findings suggest that optimized perturbations exploit a coupling between the perturbed input and the model's computation.
    \item We propose SmoothShift, a lightweight, training-free defense that combines local smoothing and spatial transformations to disrupt optimized perturbation patterns and their computational alignment while preserving task-relevant visual information.
    \item We demonstrate the effectiveness of SmoothShift across LVLMs and robustly fine-tuned CLIP models on captioning, visual question answering, and classification tasks. Further evaluations examine its robustness under defense-aware attacks and attack transfer across defense configurations.
\end{itemize}

\section{Preliminaries and Related Work}

\subsection{Preliminaries}
\label{sec:problem_formulation}

Let $\mathcal F$ denote an LVLM that generates a response from an image $u$ and a query $q$. We consider visual attacks that modify the image while keeping the query and model parameters fixed. For a clean image $x\in[0,1]^{C\times H\times W}$, the feasible set under an $\ell_\infty$ perturbation budget $\epsilon$ is
\begin{equation}
\mathcal B_\infty(x,\epsilon)
=\{u\in[0,1]^{C\times H\times W}:\|u-x\|_\infty\leq\epsilon\},
\end{equation}
where $C$, $H$, and $W$ denote the number of channels, height, and width.

Let $\mathcal L(u;q,y)$ denote the model's response loss for answer $y$. We define the attack objective $\ell(u;q,y)$ as $\mathcal L(u;q,y)$ for untargeted attacks, where $y$ is a reference answer, and as $-\mathcal L(u;q,y)$ for targeted attacks, where $y$ is the target answer. A defense-oblivious attack optimizes the image against the original model,
\begin{equation}
x^{\mathrm{adv}}\in
\arg\max_{u\in\mathcal B_\infty(x,\epsilon)}\ell(u;q,y).
\end{equation}

At inference time, an input transformation $\Phi_\theta$ with configuration $\theta$ is applied before the model generates its response,
\begin{equation}
\hat r=\mathcal F\bigl(\Phi_\theta(u),q\bigr).
\end{equation}
The defense processes both clean and adversarial inputs through the same pipeline.

\subsection{Related Work}

\textbf{Visual Adversarial Attacks on LVLMs.} Research on visual attacks against LVLMs has evolved from vulnerability assessment to increasingly transferable and query-agnostic attack formulations. Early studies established that imperceptible image-space perturbations can severely impair classification, captioning, and VQA across contemporary multimodal models~\cite{Schlarmann2023MMFM,cui2024robustness}.  Subsequent work improves transfer across models, prompts, and downstream tasks by targeting visual representations~\cite{Zhao2023AttackVLM,Wang2024VTAttack,Mei2026VEAttack}, optimizing across prompts or queries~\cite{Luo2024CroPA,Zhang2025QAVA}, or exploiting cross-modal semantic chains~\cite{Xie2025Chain}. Existing research has focused primarily on designing stronger attack objectives and achieving broader generalization. However, the mechanism of iterative optimization of these attacks remains less understood. Our work analyzes these optimization-induced properties and exploits their fragility through a lightweight test-time defense.

\paragraph{Adversarial Defenses.} Existing defenses against visual attacks on LVLMs can be broadly categorized into training-based and test-time approaches. Training-based approaches improve intrinsic robustness through text-guided contrastive adversarial training~\cite{Mao2023}, representation-preserving adversarial fine-tuning~\cite{Schlarmann2024,WangSibo2024}, or adversarial vision--language pretraining and instruction tuning~\cite{WangZeyu2025DVD}. Although effective, these approaches incur substantial training costs and may compromise the generalization ability of the models. Input-level test-time defenses process images without updating the model. Common techniques include quantization, JPEG compression, total variation minimization~\cite{guo2018countering}, and randomized transformations~\cite{raff2019barrage}. Energy-Guided Test-Time Transformation (ET3) minimizes an energy function derived from model logits to refine the input at inference time~\cite{Mirza2026ET3}. Purification-based defenses use diffusion models~\cite{Nie2022DiffPure,LiXiao2025ADBM} or LVLM-specific mechanisms~\cite{zhang2025f3}. Beyond input processing, test-time defenses intervene through prompt adaptation~\cite{WangXin2025TAPT,Sheng2025RTPT}, adversarial detection and adaptation~\cite{LiZhiwei2025TTP}, spectral rectification~\cite{Nie2026CSR}, partial-perception supervision~\cite{zhou2025dps}, or test-time counterattacks~\cite{Xing2025TTC}.

SmoothShift is an input-level test-time defense with a different design principle. Rather than estimating a clean counterpart, detecting adversarial inputs, or correcting their downstream effects, it directly disrupts properties induced by fine-grained adversarial optimization.

\section{Empirical Observations and Motivation}
\label{sec:empirical_motivation}

We first examine the sensitivity of visual adversarial attacks to optimization settings, then investigate their response to spatial rearrangement.

\subsection{Optimization Sensitivity and Local Smoothing}
\label{sec:optimization_sensitivity}

Figure~\ref{fig:motivation}(a) examines APGD~\cite{CroceHein2020AutoAttack} on COCO and TextVQA by varying the number of iterations $K$ and the minimum step-size setting $\eta_{\min}$. More iterations and smaller minimum step sizes strengthen attacks in the evaluated settings, as reflected by lower task scores. These results indicate that attack effectiveness depends on both the number of optimization iterations and the precision of perturbation refinement.

Figure~\ref{fig:motivation}(b) shows that clean and adversarial images are visually similar. Despite these subtle input differences, their representation discrepancies grow through successive layers of the vision encoder and persist in the language model, as shown in Figure~\ref{fig:motivation}(c). Together with the optimization sensitivity in Figure~\ref{fig:motivation}(a), this observation suggests that finely optimized input changes can accumulate into substantial deviations during model computation. This motivates local smoothing, which averages neighboring pixel values to disrupt fine-grained perturbation patterns while preserving task-relevant visual semantics.

\subsection{Spatial Alignment Sensitivity}
\label{sec:alignment_sensitivity}

Based on the observed optimization sensitivity and growing representation discrepancies, we hypothesize that finely optimized perturbations exploit a precise coupling between the perturbed input and the model's computation. We test this hypothesis using cyclic shifts, which rearrange pixels without changing their values. Figure~\ref{fig:motivation}(b) shows that both three- and fourteen-pixel shifts largely preserve image semantics, yet only the three-pixel shift restores the correct answer. The fourteen-pixel shift matches the patch width of LLaVA-1.5's vision encoder, preserving within-patch pixel alignment while relocating patches on the visual-token grid. Its limited mitigation suggests that changing token positions alone is less effective than disrupting within-patch alignment. Consistently, Figure~\ref{fig:motivation}(c) shows that the three-pixel shift more effectively reduces clean--adversarial representation discrepancies.

Figure~\ref{fig:motivation}(d) extends this analysis to LLaVA-1.5-7B, Qwen3.5-4B, and InternVL3.5-4B. Most shifts that are not multiples of the patch width substantially weaken attacks, whereas adversarial accuracy exhibits pronounced local minima at one and two patch widths. Clean accuracy remains comparatively stable across shifts. This consistent pattern supports our hypothesis that adversarial perturbations exploit a coupling between within-patch input patterns and the model's computation, which can be weakened by disrupting their alignment.

Together, these observations motivate a pipeline that combines local smoothing with spatial transformations. Smoothing modifies local pixel relationships, while spatial transformations alter the arrangement for which a perturbation was optimized. A mathematical analysis of this coupling is provided in the Supplementary material.

\section{SmoothShift}
\label{sec:smooth-shift}

SmoothShift combines local smoothing with a spatial transformation to disrupt finely optimized perturbation patterns. Both operations can be implemented using a variety of operators, allowing flexibility in the choice of smoothing and spatial transformations. The selected operators are applied sequentially as follows.
\begin{equation}
\Phi_{\theta}(u)=T_{\beta}\bigl(S_{\alpha}(u)\bigr),
\qquad \theta=(S,\alpha,T,\beta),
\label{eq:smoothshift_main}
\end{equation}
where $S$ and $T$ denote the selected smoothing and spatial operators, and $\alpha$ and $\beta$ their parameters.

\paragraph{Local smoothing.}
We sample from bilateral filtering, guided filtering, and local similarity averaging. Let $c$ index channels and $i,j$ index pixels. We write $u_{c,i}$ for a channel value and $u_{:,i}$ for the vector of all channel values at pixel $i$. Let $\mathcal N_k(i)$ be its $k\times k$ neighborhood and $p_i$ its spatial coordinates.

\emph{Bilateral filtering} smooths local pixel variations through weighted averaging, assigning larger weights to spatially closer neighbors with more similar colors.
\begin{equation}
\begin{aligned}
[S_{\mathrm{bil}}(u)]_{c,i}
&=\frac{\sum_{j\in\mathcal N_k(i)}w_{ij}u_{c,j}}
{\sum_{j\in\mathcal N_k(i)}w_{ij}},\\
w_{ij}&=\exp\!\left(-\frac{\|p_i-p_j\|_2^2}{2\sigma_{\mathrm{space}}^2}
-\frac{\|u_{:,i}-u_{:,j}\|_2^2}{2\sigma_{\mathrm{color}}^2}\right).
\end{aligned}
\label{eq:bilateral_smoothing}
\end{equation}
The spatial scale $\sigma_{\mathrm{space}}$ and color scale $\sigma_{\mathrm{color}}$ control the averaging range and sensitivity to color differences, respectively. The neighborhood weights $w_{ij}$ are shared across channels.

\emph{Guided filtering} smooths local pixel variations using a locally linear model guided by the input itself, with stronger averaging in low-variance regions and greater detail preservation in high-variance regions. Let $M_r$ denote box averaging with radius $r$, with local mean $m_c=M_r(u_c)$ and variance $v_c=\max\{M_r(u_c^2)-m_c^2,0\}$. Then
\begin{equation}
\begin{aligned}
[S_{\mathrm{guided}}(u)]_c&=M_r(a_c)\odot u_c+M_r(b_c),\\
a_c&=\frac{v_c}{v_c+\varepsilon_g},\qquad b_c=(1-a_c)m_c.
\end{aligned}
\label{eq:guided_smoothing}
\end{equation}
Here $a_c$ and $b_c$ are local linear coefficients, and $\varepsilon_g>0$ is a regularization parameter controlling smoothing strength. The symbol $\odot$ denotes element-wise multiplication. All operations except $M_r$ are element-wise.

\emph{Local similarity averaging} replaces each channel value with the mean of neighboring values within a similarity threshold $\tau$ of the center value. The operation is defined by
\begin{equation}
[S_{\mathrm{sim}}(u)]_{c,i}
=\frac{\sum_{j\in\mathcal N_k(i)}\mathbb{I}\{|u_{c,j}-u_{c,i}|\leq\tau\}\,u_{c,j}}
{\sum_{j\in\mathcal N_k(i)}\mathbb{I}\{|u_{c,j}-u_{c,i}|\leq\tau\}}.
\label{eq:similar_smoothing}
\end{equation}
Here $\mathbb{I}$ is the indicator function. The center is always included, ensuring a nonzero denominator. Unlike bilateral filtering's soft weights, this rule excludes dissimilar values from the average.

\begin{table*}[!t]
\caption{Performance on LLaVA-1.5-7B under APGD attacks. Average is the unweighted mean of the raw scores across four datasets. We use CIDEr for captioning and accuracy for VQA. The best result in each column is highlighted in bold, and the second-best is underlined.}
\label{tab:llava_main}

  \centering
  \setlength{\tabcolsep}{3pt}
    {\fontsize{8}{9.6}\selectfont\renewcommand{\arraystretch}{1.18}\begin{tabular*}{\textwidth}{@{}l@{\extracolsep{\fill}\hspace{8pt}}r@{\extracolsep{0pt}\hspace{7pt}}r@{\hspace{7pt}}r@{\extracolsep{\fill}\hspace{8pt}}r@{\extracolsep{0pt}\hspace{7pt}}r@{\hspace{7pt}}r@{\extracolsep{\fill}\hspace{8pt}}r@{\extracolsep{0pt}\hspace{7pt}}r@{\hspace{7pt}}r@{\extracolsep{\fill}\hspace{8pt}}r@{\extracolsep{0pt}\hspace{7pt}}r@{\hspace{7pt}}r@{\extracolsep{\fill}\hspace{8pt}}r@{\extracolsep{0pt}\hspace{7pt}}r@{\hspace{7pt}}r@{}}
    \toprule
      \multirow{2}{*}{Approach}
        & \multicolumn{3}{c}{COCO}
        & \multicolumn{3}{c}{Flickr30k}
        & \multicolumn{3}{c}{TextVQA}
        & \multicolumn{3}{c}{VQAv2}
        & \multicolumn{3}{c}{Average} \\
      \cmidrule(lr){2-4}
      \cmidrule(lr){5-7}
      \cmidrule(lr){8-10}
      \cmidrule(lr){11-13}
      \cmidrule(lr){14-16}
      {}
    & Clean & 2/255 & 4/255 & Clean & 2/255 & 4/255 & Clean & 2/255 & 4/255 & Clean & 2/255 & 4/255 & Clean & 2/255 & 4/255 \\
    \midrule
    LLaVA-1.5-7B & 115.13 & 31.35 & 14.95 & \underline{82.45} & 19.08 & 8.38 & \textbf{48.61} & 14.88 & 10.92 & \textbf{76.40} & 36.30 & 30.34 & \textbf{80.65} & 25.40 & 16.15 \\
    \midrule
    R-LLaVA-G & 106.89 & 89.97 & 76.54 & 62.84 & 49.53 & 38.90 & 25.26 & 18.74 & 14.94 & 68.54 & 59.44 & 50.02 & 65.88 & 54.42 & 45.10 \\
    AdPO & 115.30 & 68.90 & 47.60 & 75.90 & 38.60 & 27.90 & 35.50 & 24.20 & 17.60 & 73.60 & 52.30 & 37.60 & 75.08 & 46.00 & 32.68 \\
    \midrule
    Quantize & \textbf{116.02} & 71.14 & 36.61 & 81.87 & 42.62 & 20.25 & \underline{48.04} & 25.54 & 16.30 & 76.24 & 54.10 & 40.36 & \underline{80.54} & 48.35 & 28.38 \\
    TVM & 114.48 & 91.98 & 81.70 & 82.29 & 57.68 & 46.33 & 46.32 & \underline{37.64} & 33.60 & 76.09 & 67.34 & 61.74 & 79.80 & 63.66 & 55.84 \\
    Freq & \underline{115.35} & \underline{101.10} & \underline{91.84} & 78.45 & 64.48 & \underline{56.99} & 35.69 & 31.46 & 28.96 & 74.60 & 66.82 & 64.92 & 76.02 & 65.97 & 60.68 \\
    BaRT & 114.05 & 67.57 & 50.85 & 81.75 & 38.19 & 30.33 & 44.45 & 26.44 & 23.30 & 75.31 & 56.58 & 54.32 & 78.89 & 47.20 & 39.70 \\
    DPS & 104.16 & 29.01 & 13.99 & 74.35 & 16.63 & 8.02 & 43.84 & 15.52 & 11.98 & 70.68 & 35.48 & 30.68 & 73.26 & 24.16 & 16.17 \\
    F3-v3 & 100.72 & 73.43 & 68.85 & 65.67 & 42.15 & 38.17 & 31.58 & 23.70 & 21.20 & 68.24 & 58.98 & 58.62 & 66.55 & 49.57 & 46.71 \\
    TTC & 111.14 & 63.82 & 37.04 & 79.89 & 39.88 & 22.72 & 47.30 & 24.58 & 15.20 & 75.40 & 51.38 & 41.60 & 78.43 & 44.92 & 29.14 \\
    CSR & 112.33 & 63.96 & 71.27 & 80.26 & 41.01 & 43.25 & 46.86 & 22.82 & 24.18 & 75.56 & 57.12 & 58.50 & 78.75 & 46.23 & 49.30 \\
    DBD & 114.86 & 38.37 & 19.36 & \textbf{82.74} & 22.99 & 11.95 & 47.06 & 15.06 & 11.64 & \underline{76.26} & 37.38 & 31.58 & 80.23 & 28.45 & 18.63 \\
    AOM & 111.98 & 96.05 & 88.42 & 80.69 & \underline{64.88} & 56.19 & 46.89 & 37.62 & \underline{34.14} & 75.70 & \underline{71.08} & \underline{67.14} & 78.82 & \underline{67.41} & \underline{61.47} \\

    ET3 & 113.36 & 45.87 & 23.90 & 82.36 & 26.41 & 12.79 & 47.65 & 19.54 & 14.38 & 75.74 & 43.96 & 36.94 & 79.78 & 33.95 & 22.00 \\
    \midrule
    SmoothShift & 114.25 & \textbf{108.08} & \textbf{101.94} & 80.14 & \textbf{70.52} & \textbf{66.18} & 45.78 & \textbf{43.76} & \textbf{42.42} & 75.71 & \textbf{71.66} & \textbf{69.78} & 78.97 & \textbf{73.50} & \textbf{70.08} \\
    \bottomrule
    \end{tabular*}}

\end{table*}

\paragraph{Spatial transformation.}
We apply cyclic translation, rotation, or scaling to the smoothed input $v=S_{\alpha}(u)$ to alter pixel alignment with the vision encoder's patch grid.

\emph{Cyclic translation} shifts pixels vertically by $s_h$ and horizontally by $s_w$ with wraparound. For row and column indices $h,w$,
\begin{equation}
[T_{\mathrm{roll}}(v)]_{c,h,w}
=v_{c,(h-s_h)\bmod H,(w-s_w)\bmod W}.
\label{eq:translation_main}
\end{equation}
It preserves pixel values and image dimensions without interpolation, while changing their alignment with the encoder's patch grid.

\emph{Rotation} rotates the smoothed image around its center $o$, changing pixel alignment with the encoder's patch grid. For an output pixel at coordinates $p$, we sample the corresponding input location using
\begin{equation}
[T_{\mathrm{rot},\varphi}(v)]_{c,p}
=\mathcal B\!\left(v_c,\,o+R_{\varphi}(p-o)\right).
\label{eq:rotation_main}
\end{equation}
Here $R_{\varphi}$ is the rotation matrix with angle $\varphi$ used to map output coordinates to input coordinates, and $\mathcal B$ denotes bilinear interpolation with zero padding.

\emph{Scaling} similarly uses a centered coordinate map
\begin{equation}
[T_{\mathrm{scale},\rho}(v)]_{c,p}
=\mathcal B\!\left(v_c,\,o+\frac{p-o}{\rho}\right).
\label{eq:scaling_main}
\end{equation}
A factor $\rho<1$ shrinks image content on the fixed-size canvas, whereas $\rho>1$ enlarges it. These spatial operations complement smoothing by altering the input arrangement for which the perturbation was optimized.

Additional visualizations in the Supplementary material show the effects of different operator combinations on image appearance and illustrate representation recovery after defense.

\section{Experiments}

We evaluate SmoothShift's robustness across models, datasets, and attacks, and analyze its components and sensitivity to defense-aware optimization.

\subsection{Experimental Setup}
\label{sec:experimental_setup}

\textbf{Threat Model.}
We consider $L_\infty$ image-space attacks that modify only the visual input, while keeping the text query and model parameters fixed. In the main evaluation, the attacker has full white-box access to the undefended LVLM and ground-truth labels. Unless otherwise specified, adversarial examples are crafted on the base model and test-time defense is applied only at inference.

\paragraph{Attacks.} We use APGD~\cite{CroceHein2020AutoAttack,liu2026adpo} as the primary attack and further evaluate larger perturbation budgets and targeted attacks. To cover diverse attack mechanisms, we include white-box attacks FGSM~\cite{goodfellow2015explaining} and PGD~\cite{madry2018towards}, distance-regularized attacks C\&W~\cite{carlini2017towards} and EAD~\cite{chen2018ead}, the score-based black-box attack SPSA~\cite{uesato2018adversarial}, the vision-encoder attack VEAttack~\cite{Mei2026VEAttack}, and the inference-cost attack Verbose~\cite{gao2024inducing}.


%

\paragraph{Datasets and Metrics.}
We evaluate SmoothShift on both image captioning and visual question answering tasks~\cite{liu2026adpo}. We evaluate image captioning on COCO~\cite{lin2014microsoft} and Flickr30k~\cite{plummer2015flickr30k} using CIDEr~\cite{vedantam2015cider}, and visual question answering on VQAv2~\cite{goyal2017making} and TextVQA~\cite{singh2019towards} using standard VQA accuracy~\cite{antol2015vqa}. For each dataset, we use the same task-specific metric to evaluate clean performance, adversarial performance, and defended performance. Higher values indicate better task performance.

\begin{table*}[!t]
\centering
\begin{minipage}[t]{\columnwidth}
\vspace{0pt}
\caption{Results on Qwen3.5-4B under clean and adversarial settings. The best result in each column is highlighted in bold, and the second-best result is underlined.}
\label{tab:qwen}

    \setlength{\tabcolsep}{3pt}
  \centering
    {\fontsize{7}{8.4}\selectfont \begin{tabular*}{\columnwidth}{@{\extracolsep{\fill}}c@{\hspace{7pt}}r@{\hspace{7pt}}r@{\hspace{7pt}}r@{\hspace{10pt}}r@{\hspace{7pt}}r@{\hspace{7pt}}r@{}}
    \toprule
    \multirow{2}[2]{*}{Approach} 
        & \multicolumn{3}{c}{TextVQA}
        & \multicolumn{3}{c}{VQAv2} \\
      \cmidrule(lr){2-4}
      \cmidrule(lr){5-7}   
& Clean & 2/255 & 4/255 & Clean & 2/255 & 4/255 \\
    \midrule
    Qwen3.5-4B & \textbf{71.10} & 10.74 & 4.36 & \textbf{80.44} & 22.20 & 14.58 \\
    \midrule
    Quantize & 70.36 & 24.34 & 11.52 & 79.91 & 39.58 & 26.78 \\
    TVM & 68.38 & 54.34 & \underline{50.34} & 79.11 & 68.66 & 64.44 \\
    Freq & 54.91 & 47.88 & 45.04 & 78.30 & 71.22 & 69.90 \\
    BaRT & 65.67 & 34.38 & 32.68 & 78.73 & 56.54 & 50.78 \\
    DPS & 61.68 & 13.16 & 6.72 & 72.86 & 27.34 & 24.04 \\
    F3-v3 & 36.94 & 24.04 & 22.26 & 66.08 & 55.38 & 54.84 \\
    TTC & 66.92 & 24.46 & 9.16 & 78.10 & 40.12 & 25.44 \\
    CSR & 65.74 & 11.60 & 4.88 & 78.06 & 22.92 & 15.04 \\
    DBD & 66.44 & 13.54 & 5.38 & 76.78 & 27.92 & 22.30 \\
    AOM & 63.56 & \underline{54.90} & 45.58 & 76.94 & \underline{74.06} & \underline{71.58} \\

    ET3 & \underline{70.63} & 11.42 & 4.56 & \underline{80.42} & 24.26 & 16.30 \\
    SmoothShift & 66.88 & \textbf{58.86} & \textbf{56.16} & 79.07 & \textbf{74.08} & \textbf{73.56} \\
    \bottomrule
    
\end{tabular*}}

\end{minipage}
\hfill
\begin{minipage}[t]{\columnwidth}
\vspace{0pt}
\caption{Results on InternVL3.5-4B under clean and adversarial settings. The best result in each column is highlighted in bold, and the second-best result is underlined.}
\label{tab:internvl}

    \setlength{\tabcolsep}{3pt}
  \centering
    {\fontsize{7}{8.4}\selectfont \begin{tabular*}{\columnwidth}{@{\extracolsep{\fill}}c@{\hspace{7pt}}r@{\hspace{7pt}}r@{\hspace{7pt}}r@{\hspace{10pt}}r@{\hspace{7pt}}r@{\hspace{7pt}}r@{}}
    \toprule
    \multirow{2}[2]{*}{Approach} 
        & \multicolumn{3}{c}{TextVQA}
        & \multicolumn{3}{c}{VQAv2} \\
      \cmidrule(lr){2-4}
      \cmidrule(lr){5-7}   
& Clean & 2/255 & 4/255 & Clean & 2/255 & 4/255 \\
    \midrule
    InternVL3.5-4B & \underline{76.47} & 26.66 & 18.28 & \textbf{75.82} & 39.34 & 34.14 \\
    \midrule
    Quantize & 76.40 & 40.30 & 25.08 & 75.33 & 52.54 & 40.92 \\
    TVM & 75.73 & \underline{61.00} & \underline{52.88} & 74.96 & \underline{69.02} & \underline{65.32} \\
    Freq & 69.25 & 53.66 & 46.96 & 73.63 & 65.44 & 63.02 \\
    BaRT & 74.80 & 47.08 & 40.68 & 74.13 & 58.42 & 53.14 \\
    DPS & 61.69 & 23.92 & 17.56 & 65.49 & 34.48 & 31.30 \\
    F3-v3 & 51.86 & 34.84 & 33.72 & 64.20 & 53.24 & 53.42 \\
    TTC & 71.78 & 41.90 & 26.88 & 72.40 & 55.14 & 41.50 \\
    CSR & 74.10 & 28.04 & 18.22 & 75.16 & 41.38 & 34.60 \\
    DBD & 72.09 & 26.04 & 15.96 & 72.69 & 40.70 & 34.54 \\
    AOM & 73.92 & 57.28 & 46.10 & 75.28 & 64.64 & 56.88 \\

    ET3 & \textbf{76.55} & 26.76 & 17.84 & \underline{75.75} & 40.22 & 34.12 \\
    SmoothShift & 74.89 & \textbf{69.18} & \textbf{67.20} & 74.82 & \textbf{72.62} & \textbf{71.72} \\
    \bottomrule
    
\end{tabular*}}

\end{minipage}
\end{table*}

\paragraph{Implementation Details.}
For each image, we uniformly sample a smoothing operator and a spatial transformation, along with their parameters, from the candidate pools provided in the Supplementary material. We apply the sampled smoothing and spatial transformation sequentially before inference, using the same configuration for the image's clean and adversarial versions across all attack rounds.

We use the full evaluation sets for clean performance. Adversarial evaluation involves multiple rounds of iterative optimization and autoregressive generation, making it computationally expensive. Following~\cite{liu2026adpo}, we randomly sample 500 images per dataset and run up to five APGD rounds per image using different reference labels, with 100 updates per round. We retain the per-image minimum score and apply early stopping.

\begin{figure}[t]
    \centering
    \includegraphics[width=\linewidth]{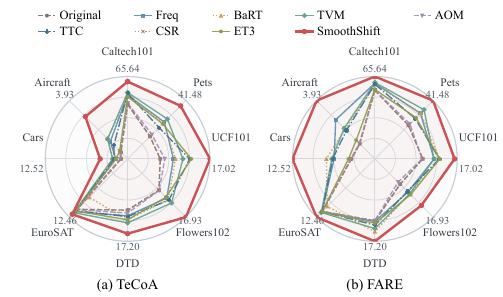}
    \caption{Adversarial accuracy on TeCoA and FARE under PGD-100 at $4/255$. Each dataset axis uses an independent scale. Outer labels show its maximum.}
    \label{fig:clip_radar}
\end{figure}

\paragraph{Robust CLIP evaluation.}
We also evaluate two robustly fine-tuned CLIP ViT-B/32 models. TeCoA uses text-guided contrastive adversarial training~\cite{Mao2023}. FARE preserves pretrained visual embeddings during unsupervised adversarial fine-tuning~\cite{Schlarmann2024}. Both checkpoints use a training perturbation budget of $4/255$. We measure zero-shot classification accuracy on Caltech101, Oxford-IIIT Pets, UCF101, Flowers102, DTD, EuroSAT, Stanford Cars, and FGVC Aircraft. Clean and adversarial evaluation use the full test splits. Attacks use defense-oblivious PGD-100 at an $\ell_\infty$ budget of $4/255$. The full protocol is provided in the Supplementary material.

\paragraph{Baseline Methods.}
 We compare SmoothShift with both training-based and test-time defenses. Training-based baselines include R-LLaVA-G, a robust LLaVA variant~\cite{cui2024robustness}, and AdPO, which improves LVLM robustness through preference optimization~\cite{liu2026adpo}. Test-time input transformations include TVM and pixel quantization~\cite{guo2018countering}, frequency compression, denoted as Freq, and the randomized transformation defense BaRT~\cite{raff2019barrage}. We further include recent vision-language test-time defenses, namely DPS~\cite{zhou2025dps}, F3-v3~\cite{zhang2025f3}, TTC~\cite{Xing2025TTC}, CSR~\cite{Nie2026CSR}, DBD~\cite{liu2026dbd}, AOM~\cite{tong2025aom}, and ET3~\cite{Mirza2026ET3}.

We use recommended baseline configurations and identical data, prompts, generation settings, adversarial inputs, and metrics for defense-oblivious evaluation. Configuration details are provided in the Supplementary material.

\begin{figure}[t]
    \centering
    \includegraphics[width=\linewidth]{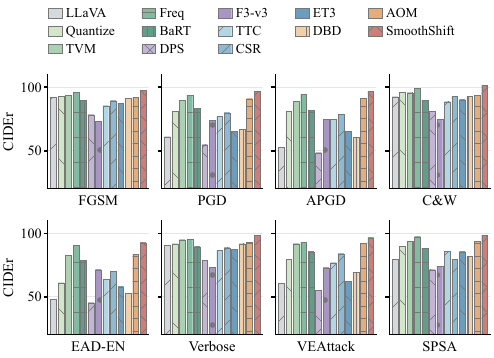}
    \caption{Mean CIDEr over COCO and Flickr30k under eight defense-oblivious attacks on LLaVA-1.5-7B. Each subplot compares 13 methods. Higher is better. All subplots share a truncated y-axis range.}
    \label{fig:attack_comparison}
\end{figure}

\begin{table}[t]
\caption{Robustness under larger perturbation budgets. Best and second-best scores are bold and underlined.}
\label{tab:large_eps}
\centering
{\fontsize{7}{8.4}\selectfont
\setlength{\tabcolsep}{0pt}
\renewcommand{\arraystretch}{1.1}
\begin{tabular*}{\columnwidth}{@{\extracolsep{\fill}}c@{\hspace{1pt}}rr@{\hspace{2pt}}rr@{\hspace{2pt}}rr@{\hspace{2pt}}rr@{}}
\toprule
 & \multicolumn{2}{c}{COCO} & \multicolumn{2}{c}{Flickr30k} & \multicolumn{2}{c}{TextVQA} & \multicolumn{2}{c}{VQAv2} \\
\cmidrule(lr){2-3}\cmidrule(lr){4-5}\cmidrule(lr){6-7}\cmidrule(l){8-9}
Approach & 16/255 & 32/255 & 16/255 & 32/255 & 16/255 & 32/255 & 16/255 & 32/255 \\
\midrule
LLaVA & 2.66 & 1.70 & 1.18 & 0.68 & 4.06 & 2.20 & 17.30 & 13.88 \\
R-LLaVA-G & 14.66 & 2.28 & 6.69 & 0.82 & 5.02 & 2.42 & 26.18 & 20.62 \\
\midrule
Quantize & 2.85 & 1.66 & 1.28 & 0.69 & 4.06 & 2.20 & 27.80 & 13.88 \\
TVM & 41.55 & 2.71 & 20.37 & 1.29 & 21.46 & 4.00 & 39.42 & 14.60 \\
Freq & \underline{73.05} & \underline{57.91} & \underline{44.98} & \underline{32.89} & \underline{23.10} & \underline{20.68} & \underline{55.40} & \underline{46.76} \\
BaRT & 18.68 & 7.12 & 9.03 & 4.29 & 13.62 & 8.98 & 36.00 & 17.56 \\
DPS & 2.83 & 1.67 & 0.97 & 0.56 & 4.60 & 2.48 & 20.40 & 15.96 \\
F3-v3 & 40.99 & 13.87 & 22.62 & 7.99 & 13.66 & 9.34 & 50.34 & 34.78 \\
TTC & 3.91 & 1.91 & 1.57 & 0.71 & 4.12 & 2.20 & 18.44 & 14.08 \\
CSR & 38.73 & 8.73 & 24.72 & 5.24 & 13.92 & 5.00 & 41.46 & 23.96 \\
DBD & 4.91 & 2.28 & 2.08 & 0.98 & 4.32 & 2.20 & 18.22 & 15.14 \\
AOM & 12.28 & 2.17 & 5.15 & 0.70 & 6.16 & 2.32 & 21.68 & 14.08 \\
ET3 & 3.54 & 1.97 & 1.57 & 0.72 & 4.12 & 2.20 & 18.84 & 13.90 \\
\addlinespace[2pt]
SmoothShift & \textbf{83.47} & \textbf{61.88} & \textbf{49.08} & \textbf{38.17} & \textbf{34.32} & \textbf{24.18} & \textbf{61.26} & \textbf{46.78} \\
\bottomrule
\end{tabular*}}
\end{table}

\subsection{Main Results}

\paragraph{Performance on LLaVA.}
Table~\ref{tab:llava_main} compares methods on LLaVA-1.5-7B across four datasets. SmoothShift achieves the highest defended score in every dataset--budget pair at $\epsilon=2/255$ and $4/255$. At $\epsilon=4/255$, it reaches 101.94 CIDEr on COCO and 42.42\% accuracy on TextVQA, compared with 14.95 and 10.92\% without defense. These results exceed the strongest competing defense in the respective columns by 10.10 CIDEr and 8.28 percentage points. Relative to the base model, clean CIDEr decreases by 0.88 on COCO and 2.31 on Flickr30k, while clean accuracy decreases by 2.83 percentage points on TextVQA and 0.69 on VQAv2.

\paragraph{Performance across LVLMs.}
\label{sec:additional_lvlms}
Tables~\ref{tab:qwen} and~\ref{tab:internvl} report results on Qwen3.5-4B~\cite{qwen35} and InternVL3.5-4B~\cite{wang2025internvl35}, respectively. SmoothShift achieves the highest defended accuracy on both datasets at both perturbation budgets for both models. At $\epsilon=4/255$, it reaches 56.16\% on TextVQA and 73.56\% on VQAv2 with Qwen, exceeding the strongest competing defense by 5.82 and 1.98 percentage points. On InternVL, the corresponding accuracies are 67.20\% and 71.72\%, with gains of 14.32 and 5.18 points. Clean accuracy decreases by 4.22 and 1.37 points on Qwen, and by 1.58 and 1.00 points on InternVL.

\paragraph{Performance on robust CLIP models.}
Figure~\ref{fig:clip_radar} reports results on the two robust CLIP models under the protocol above. SmoothShift achieves the highest adversarial accuracy among the compared methods on all eight datasets for both TeCoA and FARE. These results extend the observed benefits to robust CLIP classifiers. Detailed per-dataset results are provided in the Supplementary material.

\paragraph{Performance across attacks.}
Figure~\ref{fig:attack_comparison} compares 13 methods under eight defense-oblivious attacks using mean CIDEr over COCO and Flickr30k. SmoothShift achieves the highest captioning score under all eight attacks. These results show benefits across attack procedures beyond APGD. Full numerical results, VQA comparisons, targeted-attack evaluations, and detailed attack configurations are provided in the Supplementary material.

\paragraph{Larger perturbation budgets.}
Table~\ref{tab:large_eps} reports defense performance under APGD attacks with larger perturbation budgets of $16/255$ and $32/255$. SmoothShift achieves the highest scores on all four datasets at both budgets. Although performance decreases as the budget increases, SmoothShift retains substantial gains over the undefended model.

%

\subsection{Ablation Studies}
\label{sec:operator_ablation}
\label{sec:transformation_sensitivity}

We analyze how components and their parameters affect robustness and clean performance.

\begin{table}[t]
\centering
\caption{Component ablation of SmoothShift on LLaVA-1.5-7B. Best and second-best scores are bold and underlined.}
\label{tab:component_ablation}
{\fontsize{7}{8.4}\selectfont
\setlength{\tabcolsep}{2pt}
\renewcommand{\arraystretch}{1.12}
\begin{tabular*}{\columnwidth}{@{\extracolsep{\fill}}crrr@{\hspace{8pt}}rrr@{}}
\toprule
 & \multicolumn{3}{c}{COCO} & \multicolumn{3}{c}{Flickr30k} \\
\cmidrule(lr){2-4}\cmidrule(l){5-7}
Variant & Clean & 2/255 & 4/255 & Clean & 2/255 & 4/255 \\
\midrule
w/o spatial & \textbf{114.84} & 81.75 & 65.30 & \underline{80.84} & 50.81 & 37.77 \\
w/o smoothing & \underline{114.66} & \underline{102.91} & \underline{93.95} & \textbf{81.04} & \underline{65.89} & \underline{61.10} \\
SmoothShift & 114.25 & \textbf{108.08} & \textbf{101.94} & 80.14 & \textbf{70.52} & \textbf{66.18} \\
\midrule
\addlinespace[2pt]
 & \multicolumn{3}{c}{TextVQA} & \multicolumn{3}{c}{VQAv2} \\
\cmidrule(lr){2-4}\cmidrule(l){5-7}
Variant & Clean & 2/255 & 4/255 & Clean & 2/255 & 4/255 \\
\midrule
w/o spatial & \textbf{47.94} & 35.32 & 29.22 & \textbf{76.22} & 62.26 & 53.02 \\
w/o smoothing & \underline{46.77} & \underline{42.16} & \underline{40.38} & \underline{75.97} & \underline{68.82} & \underline{66.70} \\
SmoothShift & 45.78 & \textbf{43.76} & \textbf{42.42} & 75.71 & \textbf{71.66} & \textbf{69.78} \\
\bottomrule
\end{tabular*}}
\end{table}

\paragraph{Component contributions.}
Table~\ref{tab:component_ablation} evaluates the contributions of smoothing and spatial transformations by removing each component in turn. Spatial transformations alone outperform smoothing alone, while their combination achieves the highest robustness on all four datasets at both budgets. At $4/255$, adding smoothing to spatial transformations improves COCO CIDEr from 93.95 to 101.94 and TextVQA accuracy from 40.38\% to 42.42\%, with modest reductions in clean performance.

\begin{figure}[t]
\centering
\includegraphics[width=\linewidth]{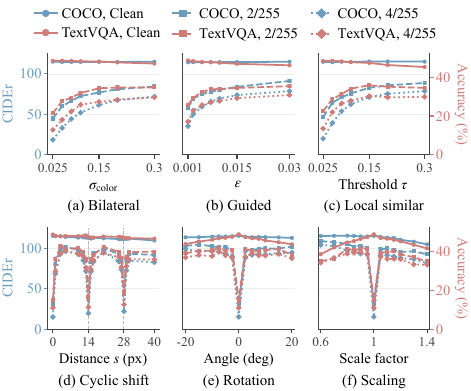}
\caption{Operator sensitivity and patch-alignment effects on COCO and TextVQA. Panel d shows reduced mitigation at shifts of 14 and 28 pixels, marked by vertical guides and corresponding to multiples of the encoder's patch size. Blue curves use the left axes, and red curves use the right axes.}
\label{fig:sensitivity_main}
\end{figure}

\paragraph{Smoothing strength.}
Figure~\ref{fig:sensitivity_main} sweeps one parameter of each standalone operator on COCO and TextVQA, with other parameters fixed. Stronger smoothing generally improves adversarial performance on both tasks, while clean COCO performance remains relatively stable and clean TextVQA accuracy declines more noticeably. This difference may reflect TextVQA's greater reliance on fine visual details for text recognition. At larger local-similarity thresholds, robustness gains saturate on COCO and diminish on TextVQA, favoring moderate smoothing.

\paragraph{Rotation and scaling.}
Both transformations improve robustness over their identity settings. Stronger transformations do not consistently yield larger gains and can reduce clean performance, particularly with aggressive downscaling on TextVQA.

\paragraph{Cyclic translation.}
Figure~\ref{fig:sensitivity_main}(d) shows that small cyclic shifts substantially improve adversarial performance while largely preserving clean performance. The gains drop sharply at 14 and 28 pixels, corresponding to one and two patch widths of LLaVA-1.5's vision encoder. This pattern supports the view that disrupting within-patch pixel alignment contributes to the robustness gains, consistent with the observations in Section~\ref{sec:alignment_sensitivity}.

Together, these single-operator analyses suggest that the observed mitigation is not specific to a single smoothing operator or spatial transformation within the evaluated defense-oblivious setting.

\subsection{Defense-Aware Evaluation}
\label{sec:cross_configuration}
\label{sec:adaptive_evaluation}

We evaluate SmoothShift against defense-aware attacks through cross-configuration transfer and mixed-gradient BPDA with EOT.

\paragraph{Cross-configuration transfer.}
We examine whether attacks optimized for one defense configuration transfer to another by evaluating
\begin{equation}
\hat r=\mathcal F\!\left(\Phi_{\theta_i}\left(x_{\theta_a}^{\mathrm{adv}}\right),q\right),
\label{eq:configuration_attack}
\end{equation}
where $x_{\theta_a}^{\mathrm{adv}}$ is optimized against configuration $\theta_a$, and $\theta_i$ is the inference configuration. For each attack-time shift $s_a$, we generate adversarial examples using one APGD round and evaluate them across inference-time shifts $s_i$, keeping all other defense parameters fixed.

\begin{figure}[t]
\centering
\includegraphics[width=\linewidth]{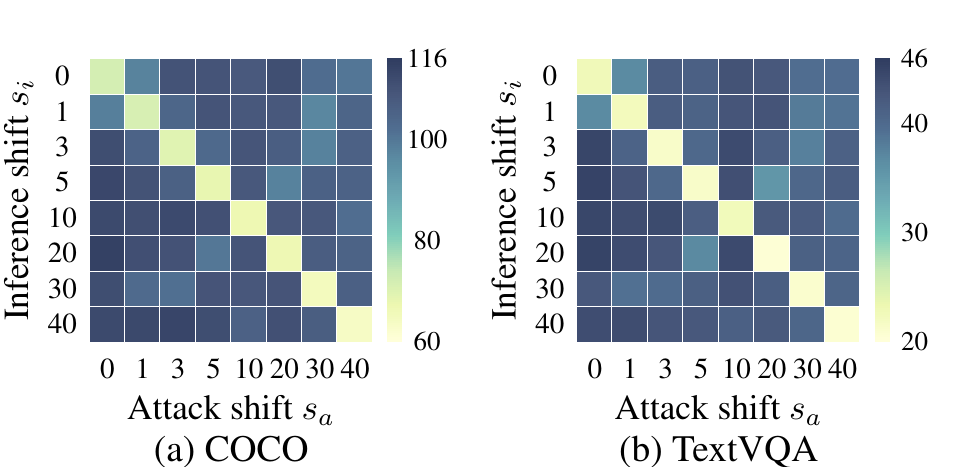}
\caption{Cross-shift transfer on COCO and TextVQA. Columns show attack shifts. Rows show inference shifts. Lower scores indicate stronger attacks.}
\label{fig:cross_configuration_main}
\end{figure}

Figure~\ref{fig:cross_configuration_main} shows that attacks are most effective when the two shifts match. Changing the inference-time shift generally weakens their effectiveness on both datasets, indicating limited cross-shift transfer. This limited transfer may benefit practical defenses when the inference configuration differs from the one used to optimize the attack.

\begin{figure}[htb]
\centering
\includegraphics[width=\linewidth]{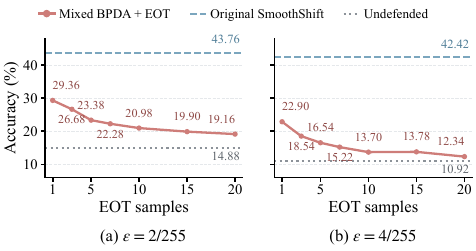}
\caption{Mixed-gradient BPDA with EOT on TextVQA. Horizontal lines show original SmoothShift and undefended performance under defense-oblivious APGD. Each setting uses 500 samples. EOT-20 reuses part of an existing attack cache, while the other settings generate fresh attacks.}
\label{fig:mixed_eot}
\end{figure}

\paragraph{Robustness under BPDA.}
We evaluate SmoothShift on TextVQA using exact gradients through differentiable operations and an approximate gradient for the hard selection in local similarity averaging. This treatment allows the attack to account for the defense operations while propagating gradients through the discrete selection. Each attack loss evaluation samples $K$ fresh configurations. EOT averages their losses to optimize against the defense distribution. An independently sampled inference configuration remains fixed across all five attack rounds. Figure~\ref{fig:mixed_eot} shows that retaining exact gradients through differentiable defense operations partially circumvents SmoothShift. Increasing the EOT sample count generally further lowers accuracy, while performance remains above the undefended baseline at both perturbation budgets. Comparisons with other defenses using whole-defense identity BPDA and EOT-20 are provided in the Supplementary material.

\section{Conclusion}

We find that visual adversarial effectiveness depends on fine-grained optimization and within-patch pixel alignment. These findings support the hypothesis that optimized perturbations exploit a coupling between the perturbed input and the model's computation. Guided by this insight, we propose SmoothShift, which combines local smoothing and spatial transformations to weaken perturbation patterns and disrupt their alignment with the encoder's patch grid. Experiments on LVLMs and robustly fine-tuned CLIP models show improved robustness with largely preserved clean performance, including gains under mixed-gradient BPDA with EOT.

\section*{LLM Usage Statement}
OpenAI ChatGPT and Codex assisted with drafting and revising text, discussing method design, and planning experiments. They also helped review related work, develop and debug code, prepare figures and tables, and organize, execute, and summarize experiments. These tools also assisted with implementation checks and analysis of experimental results. The authors remain responsible for the correctness, reproducibility, attribution, and interpretation of all submitted content.

\bibliographystyle{IEEEtran}
\bibliography{main}
\end{document}